\documentclass{article}

\usepackage{PRIMEarxiv}

\usepackage[utf8]{inputenc}
\usepackage[T1]{fontenc}
\usepackage{amsmath}
\usepackage{amssymb}
\usepackage{amsfonts}
\usepackage{graphicx}
\usepackage{booktabs}
\usepackage{url}
\usepackage[hidelinks]{hyperref}
\usepackage{float}
\usepackage{placeins}
\graphicspath{{media/}}

\title{Inferring Soil Friction Angle from Robot Foot-Ground Force Histories: A Bayesian Inverse Approach to Proprioceptive Soil Sensing}

\author{
  Dawei Xu, Zhijie Wang\thanks{Corresponding author: \texttt{zhijie.wang3@wsu.edu}} \\
  Department of Civil and Environmental Engineering \\
  Washington State University \\
  Pullman, WA 99164, USA \\
  \texttt{dawei.xu@wsu.edu, zhijie.wang3@wsu.edu}
}

\begin{document}

\maketitle

\begin{abstract}

Foot-ground interaction signals recorded by quadruped robots may enable spatially distributed, in situ characterization of soil strength. As a first step, we test whether the internal friction angle $\phi$ of cohesionless soil can be identified from the force history of a simplified rotating leg. A two-dimensional continuum model implemented with the material point method, benchmarked against measured rotating-leg force histories, generates the training data, and two Gaussian-process surrogates support Bayesian inversion of the full histories. In matched-model experiments, the framework recovers 14 off-grid friction angles with a median absolute error of approximately $0.1^\circ$ (maximum $\sim\!0.7^\circ$); the reported credible intervals contain the true value in every case. These results establish that $\phi$ is identifiable when the forward model is correctly specified, and support further development of proprioceptive soil sensing for spatially variable terrain, with applications from physics-grounded world models for robot training to post-wildfire slope assessment.

\end{abstract}

\section{Introduction}
\label{sec:intro}

Each footfall of a quadruped acts as a localized mechanical probe of the
terrain. Joint torques and leg kinematics can be used to reconstruct the
associated contact forces without dedicated foot-force sensors
(Section~\ref{sec:field}), producing a force-motion record at every step. This
raises a fundamental question: can these routinely acquired signals reveal
quantitative properties of the underlying soil? If so, locomotion could become
a source of distributed terrain measurements rather than solely a means of
transport. Such a capability would support applications from post-wildfire slope assessment to in-situ soil-parameter calibration for high-fidelity world models.

Spatial variation in near-surface soil strength affects bearing response,
trafficability, deformation, soil erosion, and slope stability \cite{castrignano20023d, jiang2022advances,Wang2025geotech,Wang2024threedim}. Conventional characterization
with cone penetrometers, shear vanes, plate-load devices, and laboratory shear
tests provides direct measurements, but requires dedicated equipment, operator
access, and separate sampling campaigns \cite{wu1996soil, budhu2010soil}. Moreover, such conventional means become prohibitive in challenging terrains due to limited accessibility. A legged robot could complement these
methods by collecting repeated interaction measurements along its path while
performing its primary mission in rough terrains. Among the relevant strength properties, the internal friction angle $\phi$
governs frictional shear resistance in cohesionless soil and influences both
load-bearing response and slope stability. Estimating $\phi$ from foot-soil
force histories would therefore connect proprioceptive terrain sensing with a
physically interpretable constitutive parameter, rather than only a terrain
class or traversability label.

\begin{figure}[!htbp]
\centering
\includegraphics[width=0.75\linewidth]{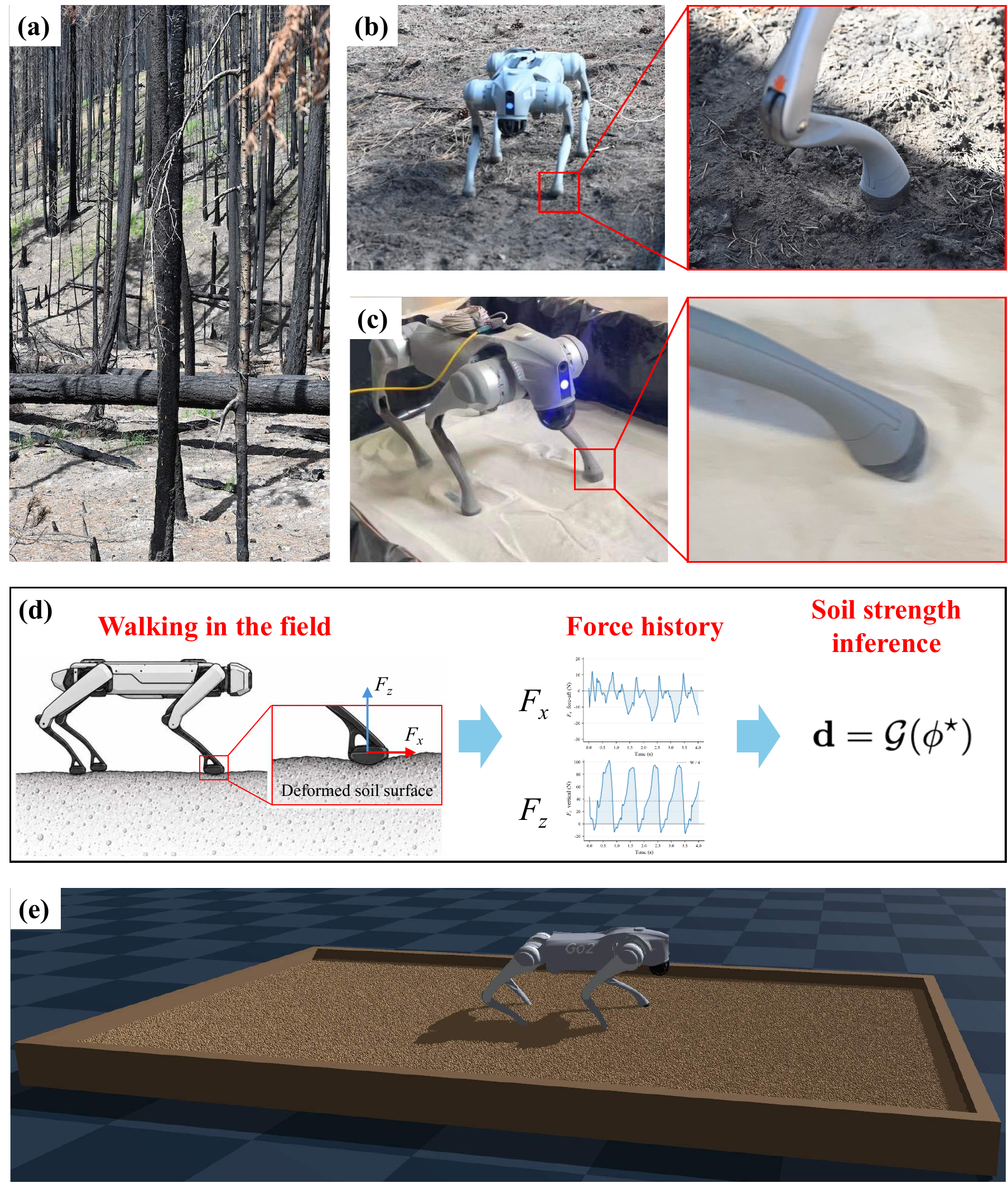}
\caption{Physics-grounded real-to-simulation-to-real framework for legged-robot
terrain understanding: (a,b)~a wildfire-burned site and field deployment on it;
(c)~laboratory characterization of physical terrain properties; (d)~the proposed inversion of robot force histories for soil-strength inference; and (e)~inferred physical
properties parameterize high-fidelity granular terrain models for robot training and redeployment (generated using the Genesis platform \cite{genesis2024}). The loop connects real-world robot–terrain interaction with physics-grounded world modeling, enabling continual refinement of terrain representation and robot behavior.}
\label{fig:workflow}
\end{figure}

This capability would serve two applications in particular. The first is the construction of high-fidelity world models for robot training. Geometric world models describe environmental appearance and structure, but contact-rich robot training also requires credible terrain mechanics. We therefore envision a physics-grounded real-to-simulation-to-real loop (Fig.~\ref{fig:workflow}): proprioceptive interaction data collected in the field are inverted for soil properties, which parameterize deformable terrain in a virtual training world. Policies trained in this mechanically grounded environment can then be redeployed, with subsequent interactions providing new data for model refinement. The second is post-wildfire hillslope assessment. Heating and the loss of vegetation and litter can alter soil structure, organic matter, wettability, hydrologic response, and near-surface mechanical resistance \cite{certini2005effects}. Subsequent rainfall can increase susceptibility to erosion, debris flows, and shallow slope failures \cite{staley2017prediction,wang2025multiscale,Wang2025part2}. Combined with topographic, hydrologic, and other geotechnical inputs, spatially distributed strength estimates could support rapid estimation of a slope's factor of safety using physics-guided machine learning models \cite{xu2027physics}, enabling fast identification of areas at elevated risk of slope instability. The present study addresses an enabling real-to-simulation step in this loop by identifying soil friction angle from leg-soil force histories.

Previous studies on soil-property inference from robot-ground interaction can be grouped into four areas. Classical terramechanics \cite{bekker1969introduction,wong1967prediction,wong2008theory} established semi-empirical relationships between soil properties and wheel or plate responses, with parameters typically obtained from dedicated bevameter or plate-load tests. Planetary-rover studies \cite{iagnemma2004online,ding2011experimental} extended these relationships to in situ estimation of cohesion, internal friction angle, and related terrain parameters, but from scalar interaction features such as wheel torque, slip, and sinkage rather than a full force history. In legged robotics, proprioceptive signals have been used for terrain classification, traversability, traction, and slip estimation \cite{hoepflinger2010haptic,wellhausen2019where,kolvenbach2019haptic}, but less attention has been given to constitutive soil parameters. Simulation-based inverse approaches \cite{coetzee2009calibration,li2013terradynamics,agarwal2021surprising} have also inferred soil properties from intrusion, rotating-leg, plate-sinkage, and excavation responses, generally using dedicated laboratory measurements. Across these areas, quantitative in-situ estimation of $\phi$ has been demonstrated mainly for wheeled platforms from scalar slip and sinkage; whether the full time-resolved force history of a single leg carries a comparably identifiable signature of $\phi$, recoverable from onboard signals, remains untested.

We propose a framework for inferring an interpretable soil-strength parameter
from proprioceptive leg-force histories in dry, cohesionless soil. A continuum forward model of an idealized rotating leg generates horizontal and vertical force histories over a range of friction angles; two Gaussian-process surrogates, one per force component, together with Bayesian inversion then recover $\phi$ from the full history. Field recordings from a quadruped robot illustrate that the required force signal is available onboard, and published rotating-leg measurements benchmark the forward model. We evaluate feasibility through off-grid, matched-model recovery tests. This study thus isolates the central identifiability question underlying robot-assisted soil characterization. This scope, for which the friction angle governs shear strength, excludes cohesive and partially saturated soils, which are left to future work.

\FloatBarrier
\section{Problem formulation and field context}
\label{sec:formulation}

Turning proprioceptive interaction signals into soil properties requires two
conditions. First, the target property must leave a distinguishable signature
in the leg--soil force history under a controlled interaction. Second, that
history must be observable from onboard robot measurements.
Section~\ref{sec:inverse-problem} formulates the first condition as an inverse
problem for friction angle, and Section~\ref{sec:field} examines the second
using field locomotion data. Separating these
questions allows the fundamental identifiability of the soil parameter to be
tested before addressing the full complexity of field deployment.

\subsection{An inverse problem for the friction angle}
\label{sec:inverse-problem}

Natural locomotion combines variations in foot trajectory, loading rate,
geometry, and terrain condition. We therefore begin with a necessary,
controlled question: when the leg geometry and motion are prescribed, can the
internal friction angle \(\phi\) of cohesionless soil be identified from the
resulting force history? For zero cohesion, the Mohr--Coulomb shear strength is
\begin{equation}
    \tau = \sigma \tan\phi,
    \label{eq:mc}
\end{equation}
where \(\sigma\) is the effective normal stress. Because \(\phi\) controls the
mobilized shear resistance, it affects both the magnitude and evolution of the
reaction force as a leg moves through soil.

To isolate that relationship from gait-level variability, we consider a rigid
leg of fixed geometry (chord \(2R\) and width \(w\)) rotating through soil with
a prescribed angular-rate profile \(\omega(\theta)\). The forward operator maps
the friction angle to the horizontal and vertical force histories,
\begin{equation}
    \mathcal{G}:\ \phi \longmapsto (F_x,F_z)(\theta),
    \label{eq:fwd}
\end{equation}
and is evaluated numerically in Section~\ref{sec:forward}. Both components and
their full evolution with \(\theta\) are retained because \(\phi\) can affect
the force magnitude and history shape.

For an observed force history
\begin{equation}
    \mathbf{d}=\mathcal{G}(\phi^\star),
\end{equation}
the inverse problem is to estimate the unknown friction angle \(\phi^\star\).
Bayesian inversion (Section~\ref{sec:inverse-method}) yields a posterior
distribution over \(\phi\). Only \(\phi\) is inferred; density and elastic
properties are held fixed, so the result is conditional on the adopted
constitutive model and nuisance parameters. Sections~\ref{sec:forward} and
\ref{sec:inverse} test this controlled identifiability through forward-model
validation, force-sensitivity analysis, and held-out parameter recovery. The
following subsection first establishes whether the required force-history
signal is available from robot proprioception.

\subsection{Proprioceptive force histories from field locomotion}
\label{sec:field}

To examine the second condition, we use field recordings to determine whether
onboard measurements provide structured, repeatable foot--soil force histories.
These data establish the sensing context and are not used as observations in
the parameter-recovery experiments.

We analyzed proprioceptive recordings from a Unitree Go2 walking at
\(0.3\,\mathrm{m\,s^{-1}}\) along a \(4\,\mathrm{m}\) dry, sandy path within a
wildfire-burned field site. Joint torque and position were recorded at approximately
\(500\,\mathrm{Hz}\). For each leg, the contact force was reconstructed from
the joint configuration $q$, actuator torque $\boldsymbol{\tau}$, and analytic
Jacobian $J(q)$ using the quasi-static virtual-work relation \cite{yang2025high}
\begin{equation}
    \widehat{\mathbf{f}}
    =-\left[J(q)^{\top}\right]^{-1}\boldsymbol{\tau}.
\end{equation}
Because gravity, link inertia, motor friction, and controller feedforward are
not explicitly compensated, $\widehat{\mathbf{f}}$ is treated as an approximate
interaction force.

\begin{figure}[!htbp]
  \centering
  \includegraphics[width=0.75\linewidth]{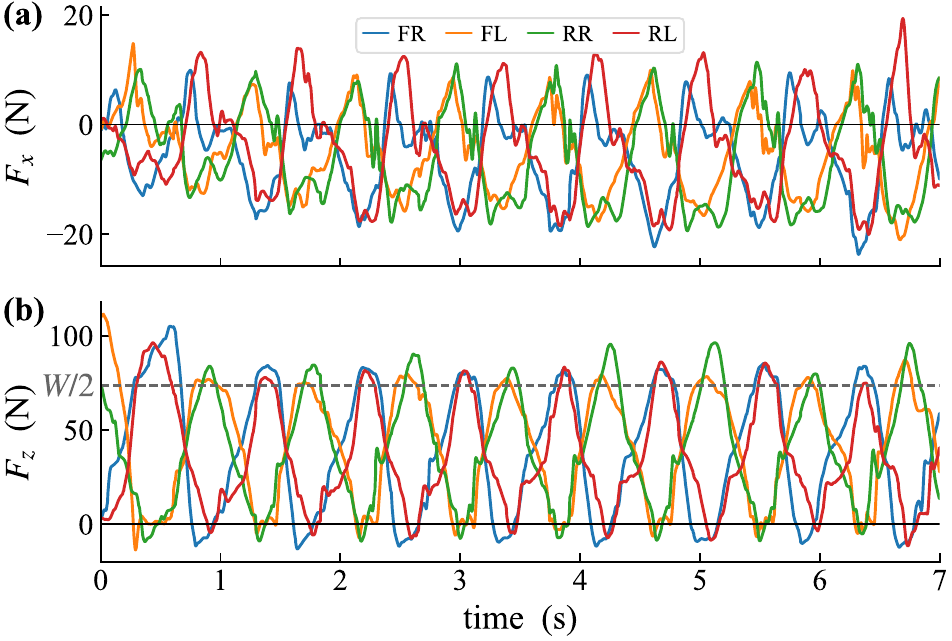}
  \caption{Proprioceptive foot-force estimates for a Unitree Go2 walking on
    sand: (a)~fore--aft force $F_x$ and (b)~vertical force $F_z$ over
    $7\,\mathrm{s}$.}
  \label{fig:field-forces}
\end{figure}

Figure~\ref{fig:field-forces} shows the reconstructed fore--aft (\(F_x\)) and
vertical (\(F_z\)) forces during a representative walk.
A stance interval is defined around each $F_z$ peak, extending from the preceding to the subsequent zero crossing of $F_z$.
Repeated stance pulses are evident for all four feet: \(F_z\) rises
to approximately one-half of the robot weight, while \(F_x\) is
predominantly negative before becoming positive near push-off.

\begin{figure}[!htbp]
  \centering
  \includegraphics[width=0.75\linewidth]{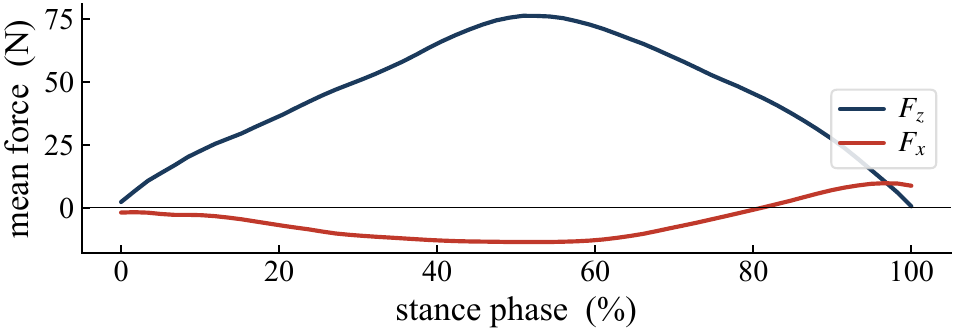}
  \caption{Stride-aligned mean stance forces from multiple stance pulses
    ($68$ pulses across four feet).}
  \label{fig:field-ensemble}
\end{figure}

All the stance pulses were resampled onto a single pair of stance force curves with the time normalized
(Fig.~\ref{fig:field-ensemble}). Their mean peaks are approximately
\(76\,\mathrm{N}\) in \(F_z\) and \(-14\,\mathrm{N}\) in \(F_x\).
Here, positive $F_x$ denotes a forward ground-reaction force and
negative $F_x$ a backward ground-reaction force, while positive $F_z$
denotes an upward ground-reaction force and negative $F_z$ a downward force.
The field data therefore demonstrate the availability
of the target signal that can be used for proprioceptive sensing.
The following controlled analyses test whether that
signal contains an identifiable signature of \(\phi\).

\FloatBarrier
\section{Forward modeling}
\label{sec:forward}

\subsection{Physics-based model and numerical implementation}
\label{sec:mpm-setup}

Leg--soil interaction involves large deformation, a moving free surface, and
repeated contact. Grain-resolved discrete-element models can capture individual
contacts but are costly for repeated parameter sweeps
\cite{coetzee2009calibration}, whereas conventional finite-element meshes may
distort under large soil motion \cite{nazem2021alternative}.
We therefore implement the continuum forward
model with the material point method (MPM), which carries material history
through a background grid and accommodates large deformation without persistent
mesh distortion \cite{soga2016trends}.

The model is two-dimensional and plane strain \cite{sulsky1994particle}.
Momentum and stress are transferred between material points and a Cartesian
background grid using quadratic B-splines and a moving-least-squares
discretization \cite{jiang2015affine,hu2018mlsmpm}.

The soil follows a hypoelastic--plastic Drucker--Prager model
\cite{drucker1952soil}. Its parameters are related to the Mohr--Coulomb cohesion
$c$ and friction angle $\phi$ along the triaxial-compression meridian
\cite{chen1988plasticity,klar2016drucker}. With tension-positive stress,
$p=\tfrac12\operatorname{tr}\boldsymbol{\sigma}$,
$\mathbf{s}=\boldsymbol{\sigma}-p\mathbf{I}$, and
$\lVert\mathbf{s}\rVert_\ast=\sqrt{\tfrac12\mathbf{s}:\mathbf{s}}$, the yield
function is
\begin{equation}
  f(\boldsymbol{\sigma}) = \lVert\mathbf{s}\rVert_\ast
  + \alpha_\phi\, p - k_c,
  \label{eq:dp-yield}
\end{equation}
with
\begin{equation}
  \alpha_\phi = \frac{2\sin\phi}{\sqrt{3}\,(3-\sin\phi)},
  \quad
  k_c = \frac{6\,c\cos\phi}{\sqrt{3}\,(3-\sin\phi)},
  \label{eq:dp-consts}
\end{equation}
where $c$ is the cohesion and $\phi$ the friction angle. Under dry, cohesionless conditions, we set $c=0$
throughout, including all three leg geometries of the benchmark in
Section~\ref{sec:verification}, and infer $\phi$ in
Section~\ref{sec:inverse}. A standard associated-flow return mapping with a
tension cutoff enforces the rate-independent yield response
\cite{simo1998computational,desouzaneto2008computational}. Particles are initialized
in a geostatic state with $\sigma_{yy}=-\rho g(h_s-y)$,
$\sigma_{xx}=K_0\sigma_{yy}$, and $K_0=1-\sin\phi$  \cite{jaky1944coefficient}.
Although $K_0$ therefore varies with $\phi$, a
companion sweep with $K_0=0.5$ changes the force library by less than $0.5\%$,
confirming that the observed sensitivity is governed by frictional strength
rather than the initial stress.

The rigid leg is represented by an analytic signed-distance function. Contact
enforces non-penetration and Coulomb friction, and nodal momentum changes give
the reaction components $F_x$ and $F_z$. The benchmark geometry, loading, and
discretization are specified next.

\subsection{Model calibration}
\label{sec:verification}

\begin{figure}[!htbp]
\centering
\includegraphics[width=0.75\columnwidth]{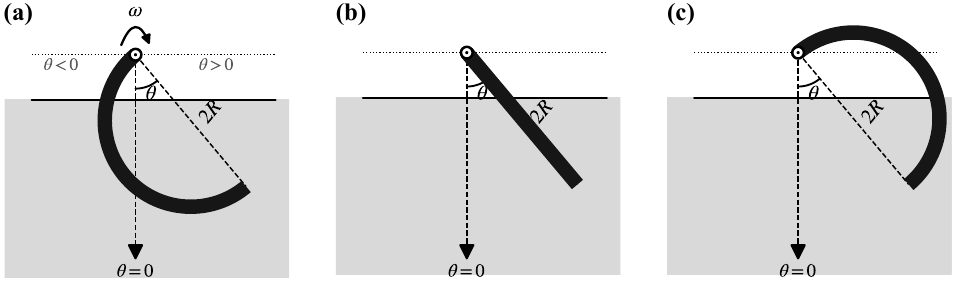}
\caption{Rotating legs in sand with different geometries: (a)~C-leg ($\kappa=+1/R$), (b)~straight leg
($\kappa=0$), and (c)~reversed-C leg ($\kappa=-1/R$). Here, $\theta$ is
measured from vertical and $2R$ is the chord length.}
\label{fig:leg-geometry}
\end{figure}

The rotating-leg experiments of Li et al.\ \cite{li2013terradynamics} are used
to benchmark the numerical forward model. The experiments considered straight,
C-shaped, and reversed-C legs rotating through dry Yuma sand at
$\omega=0.2~\mathrm{rad\,s^{-1}}$. All three legs had a chord length of
$2R=76.2~\mathrm{mm}$ and a width of $w=25.4~\mathrm{mm}$. We denote the
signed centerline curvature by $\kappa$: $\kappa=0$ for the straight leg and
$\kappa=\pm1/R$ for the semicircular legs, with the positive and negative
signs corresponding to the C and reversed-C orientations, respectively
(Fig.~\ref{fig:leg-geometry}). The measured
horizontal and vertical force histories, $F_x(\theta)$ and $F_z(\theta)$,
were digitized from Fig.~S12 of Li et al.\ \cite{li2013terradynamics}
and compared with the corresponding model predictions.
A common set of numerical and constitutive parameters was
used for all three geometries.

The benchmark configuration is summarized in
Table~\ref{tab:benchmark-config}. The friction angle and bulk density were
taken from the properties reported for the experimental medium. Mechanical parameters
not reported for the benchmark, including the elastic properties and
leg--soil interface friction, were calibrated with the experimentally measured force histories. Cohesion and dilation were
set to zero for the benchmark. A small numerical damping is applied to suppress grid-scale oscillations associated with the explicit contact calculation \cite{tran2019temporal}.
After each time step, the background-grid velocity is multiplied by $0.997$ before the
grid-to-particle transfer, corresponding to a characteristic decay time of approximately
$2.7~\mathrm{ms}$. The domain boundaries enforce non-penetration, and leg--soil contact is
applied within $2\Delta x$ of the analytic signed-distance surface. In order to represent lateral
soil escape around a finite-width leg in the two-dimensional plane-strain formulation,
we approximate this three-dimensional relief mechanism using a material-point factor $q_p$,
initialized as $q_p=1$ and updated according to
\begin{equation}
q_p^{n+1} =
\begin{cases}
\max\!\bigl(q_{\min},\,q_p^n e^{-k|\Delta\theta|}\bigr), & \text{(churned)}\\[2pt]
q_p^n, & \text{(otherwise)}
\end{cases}
\label{eq:relief}
\end{equation}
where a point is ``churned'' when $\theta^{n+1}>0$ and
$\|\mathbf{x}_p^n-\mathbf{x}_p^0\|>d_r$. We use
$k=2~\mathrm{rad}^{-1}$, $d_r=20~\mathrm{mm}$, and $q_{\min}=0.05$; $q_p$
scales the particle contributions to mass, momentum, and stress. These values
were selected once for the benchmark and retained across all geometries,
friction angles, and rotation rates. Because the correction acts only during
withdrawal, it leaves loading up to the deepest point unchanged.

\begin{table}[htbp]
\centering
\caption{Rotating-leg benchmark parameters.}
\label{tab:benchmark-config}
\footnotesize
\setlength{\tabcolsep}{4pt}
\begin{tabular}{ll}
\toprule
Parameter & Value \\
\midrule
\multicolumn{2}{l}{\textit{Experiment}}\\
\quad Leg geometry & straight, C-shaped, reversed-C \\
\quad Angular velocity, $\omega$ & $0.2~\mathrm{rad\,s^{-1}}$ \\
\quad Chord length, $2R$ / width, $w$ & $76.2$ / $25.4~\mathrm{mm}$ \\[2pt]
\multicolumn{2}{l}{\textit{Discretization}}\\
\quad Domain / soil-bed size & $0.30\times0.30$ / $0.29\times0.165~\mathrm{m}$ \\
\quad Background grid, $\Delta x$ & $128^2$, $2.34~\mathrm{mm}$ \\
\quad Material points per cell & $4$ \\
\quad Time step, $\Delta t$ & $8\times10^{-6}~\mathrm{s}$ \\[2pt]
\multicolumn{2}{l}{\textit{Contact / numerics}}\\
\quad Contact range & $2\Delta x$ \\
\quad Interface friction, $\mu_{\mathrm{leg}}$ & $0.35$ \\
\quad Grid-velocity damping & $0.997$ per step \\[2pt]
\multicolumn{2}{l}{\textit{Soil}}\\
\quad Friction angle, $\phi$ & $35^\circ$ \\
\quad Density, $\rho$ & $1650~\mathrm{kg\,m^{-3}}$ \\
\quad Young's modulus, $E$ / Poisson, $\nu$ & $20~\mathrm{MPa}$ / $0.30$ \\
\quad Cohesion, $c$ / dilation, $\psi$ & $0$ / $0^\circ$ \\
\bottomrule
\end{tabular}
\end{table}

\begin{figure}[!htbp]
\centering
\includegraphics[width=0.75\linewidth]{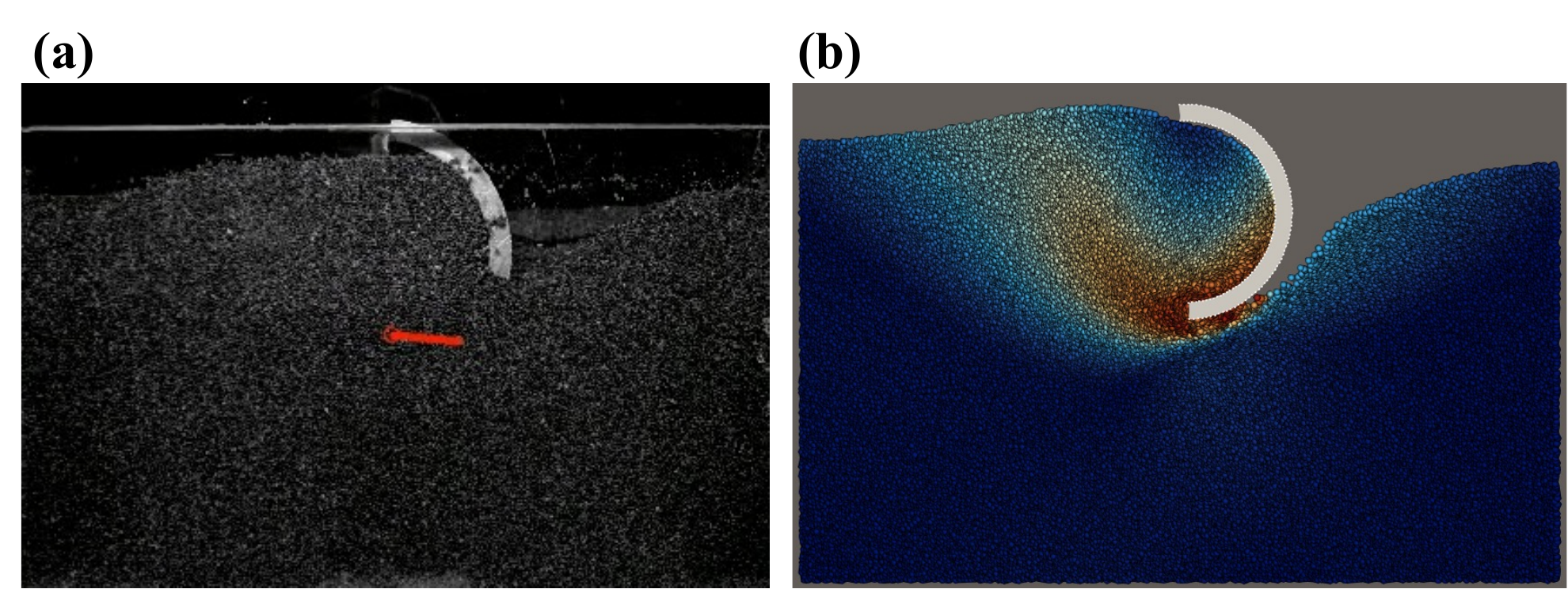}
\caption{Reversed-C leg in sand at a matched sweep angle. (a)~experiment \cite{li2013terradynamics}, where the leg is partially obscured by sand and the red arrow marks the resultant contact force vector; (b)~simulated material points colored by velocity magnitude (blue: low, red: high).}
\label{fig:comparison-rcleg}
\end{figure}

\begin{figure}[!htbp]
\centering
\includegraphics[width=0.75\columnwidth]{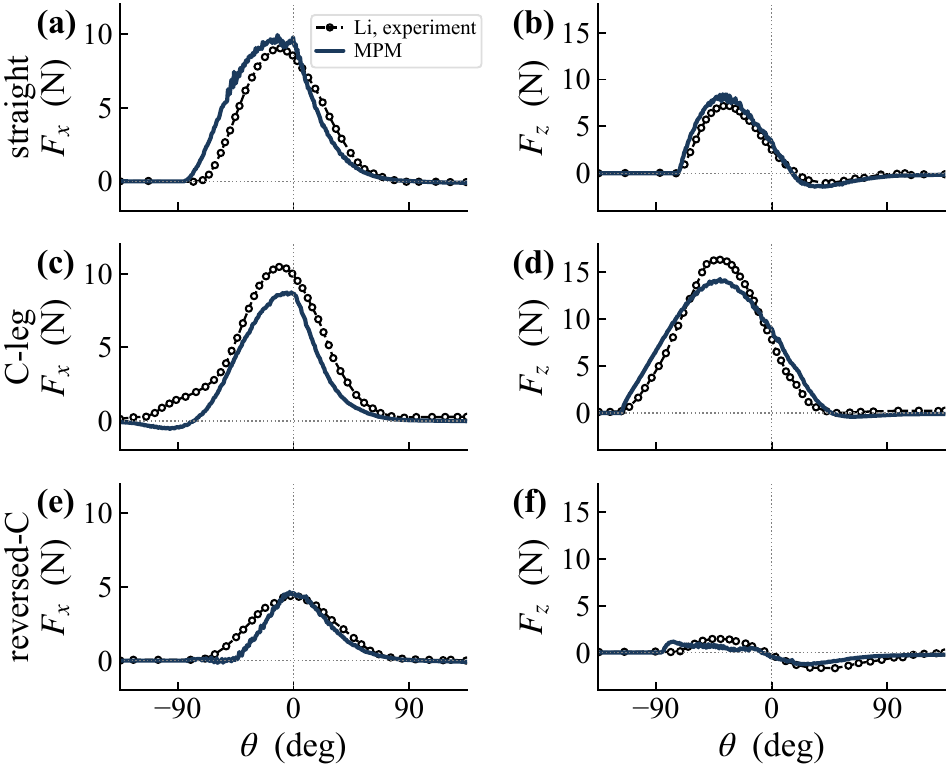}
\caption{Comparison of simulated and measured rotating-leg force histories at $\omega=0.2~\mathrm{rad\,s^{-1}}$ \cite{li2013terradynamics}. Rows show the straight, C-shaped, and reversed-C legs; columns show $F_x$ and $F_z$.}
\label{fig:verification}
\end{figure}

The simulated material-point configuration reproduces the soil deformation observed in the experiment, including the bulldozed pile that forms ahead of the reversed-C leg and motivates the relief correction above (Fig.~\ref{fig:comparison-rcleg}). Fig.~\ref{fig:verification} compares the simulated force histories with the
experimental measurements of Li et al.\ \cite{li2013terradynamics}, and
Table~\ref{tab:verification} summarizes the whole-curve agreement between the simulated and measured force histories, quantified by the normalized root-mean-square deviation (NRMSD) over $|\theta|<85^\circ$.
Compared with the model predictions reported by Li et al.\ \cite{li2013terradynamics}, the MPM forward model achieves comparable agreement for the straight and C-shaped legs and improved agreement for the more challenging reversed-C geometry. More importantly, its errors are relatively uniform across all three geometries and both force components, with NRMSD values confined to $14.9\%$--$24.8\%$. This consistency across substantially different leg curvatures indicates that the forward model captures the principal geometry-dependent features of the measured soil reaction, rather than fitting any single geometry. This uniform, geometry-independent error behavior is what makes the model suitable as the forward operator for the Gaussian-process surrogate and Bayesian inversion that follow, since the inversion relies on the model reproducing the correct force--$\phi$ relationship consistently rather than on its accuracy for one configuration. Although all three geometries achieve comparable agreement with experiment, the subsequent friction-angle library and inverse analysis use the straight leg. Its flat, uncurved contact geometry is the neutral choice: it isolates the friction-angle response from curvature-induced effects such as soil trapping. 

\begin{table}[htbp]
\centering
\caption{MPM force-history deviations from experiment over
$|\theta|<85^\circ$, normalized by the experimental peak-to-peak range
(NRMSD, \%).}
\label{tab:verification}
\setlength{\tabcolsep}{5pt}
\begin{tabular}{l cc}
\toprule
Leg & $F_z$ NRMSD (\%) & $F_x$ NRMSD (\%) \\
\midrule
Straight    & 24.8 & 14.9 \\
C-shaped    & 17.6 & 15.0 \\
Reversed-C  & 18.7 & 14.9 \\
\bottomrule
\end{tabular}
\end{table}

\subsection{Force response to the friction angle}
\label{sec:straightleg}

We next quantify the sensitivity of the simulated force response to the friction angle $\phi$. The simulations use the straight-leg configuration of Section~\ref{sec:mpm-setup}, a sweep range $\theta\in[-135^\circ,135^\circ]$, and the benchmark rotation rate $\omega=0.2~\mathrm{rad\,s^{-1}}$ (Section~\ref{sec:verification}). The friction angle is varied over
$\phi\in[25^\circ,45^\circ]$. This interval spans critical-state friction
angles reported for dry sands over a range of effective stress levels
\cite{rouse2018relation} and is adopted here as a broad, physically
plausible range for the sensitivity analysis.

\begin{figure}[H]
\centering
\includegraphics[width=0.75\linewidth]{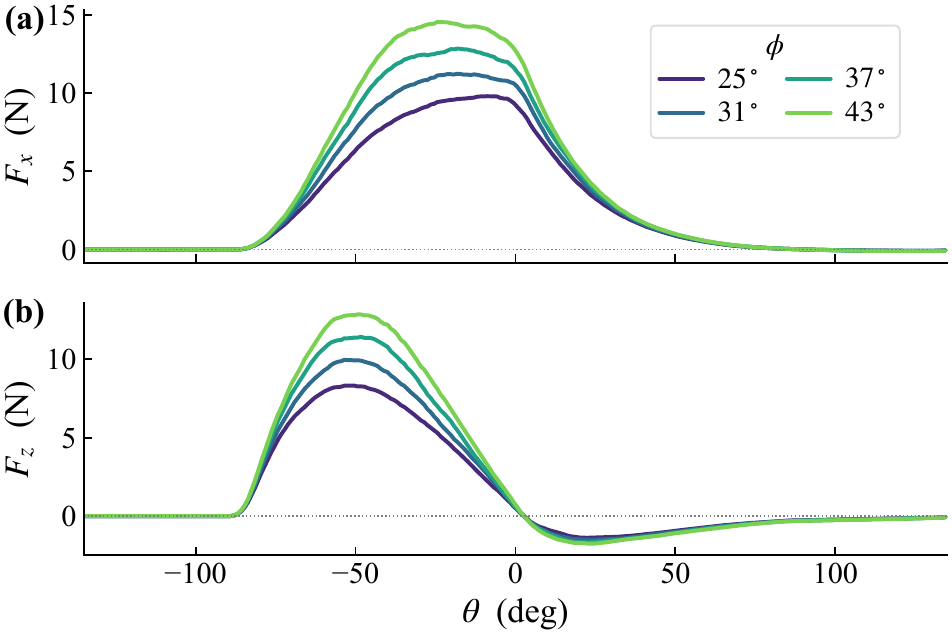}
\caption{Straight-leg force histories for $\phi=25^\circ$, $31^\circ$, $37^\circ$, and
$43^\circ$ at $\omega=0.2~\mathrm{rad\,s^{-1}}$: (a)~$F_x$ and (b)~$F_z$.}
\label{fig:phi-sensitivity}
\end{figure}

Fig.~\ref{fig:phi-sensitivity} shows the resulting force histories. The leg
engages the soil near $\theta=-85^\circ$, after which the vertical and
horizontal reaction forces increase as the leg penetrates the bed.
The vertical force $F_z$ reaches its maximum near $\theta=-50^\circ$,
whereas the horizontal force $F_x$ peaks near $\theta=0^\circ$. During
withdrawal, the forces decrease and $F_z$ eventually reverses sign. The response varies systematically with $\phi$. Over
$\phi=25^\circ$--$45^\circ$, peak $F_z$ increases from $8.6~\mathrm{N}$ to
$14.0~\mathrm{N}$ ($63\%$), and peak $F_x$ from $10.2~\mathrm{N}$ to
$15.5~\mathrm{N}$ ($53\%$). The descending and withdrawal portions also
change shape, indicating that information about $\phi$ is distributed over the
full force trajectory rather than confined to its peaks. The complete straight-leg simulation library used in the inverse analysis is
summarized in Table~\ref{tab:library}.

\clearpage

\begin{table}[htbp]
\centering
\caption{Straight-leg simulation library and off-grid test cases.}
\label{tab:library}
\begin{tabular}{ll}
\toprule
Parameter & Values \\
\midrule
Friction angle, $\phi$
    & $25^\circ$--$45^\circ$, $2^\circ$ steps (11 cases) \\
Rotation rate, $\omega$
    & $0.2~\mathrm{rad\,s^{-1}}$ (matches the benchmark) \\
Off-grid test angles
    & 14 cases, $\phi\in[25.9^\circ,43.6^\circ]$ \\
Sweep range, $\theta$
    & $[-135^\circ,135^\circ]$ \\
\bottomrule
\end{tabular}
\end{table}

\FloatBarrier
\section{Inverse analysis}
\label{sec:inverse}

Inverse analysis has been used in geotechnical engineering to identify soil
parameters by reconciling measured responses with numerical predictions
\cite{calvello2004selecting,zhang2018determination}. Bayesian formulations
extend this approach by representing parameter and model uncertainty through a
posterior distribution \cite{wang2016bayesian,zhao2024uncertainty,
murakami2023developments}. Here, the measured response is a leg-force history
and the target parameter is the soil friction angle.

\subsection{Method}
\label{sec:inverse-method}

The inverse problem is formulated as estimation of the friction angle $\phi$
from a leg-force history. For an observation vector $\mathbf{d}$, Bayes' rule gives
the posterior
\begin{equation}
  p(\phi\mid\mathbf{d}) \;\propto\; p(\mathbf{d}\mid\phi)\,p(\phi),
  \label{eq:bayes}
\end{equation}

\begin{figure}[!htbp]
  \centering
  \includegraphics[width=0.75\linewidth]{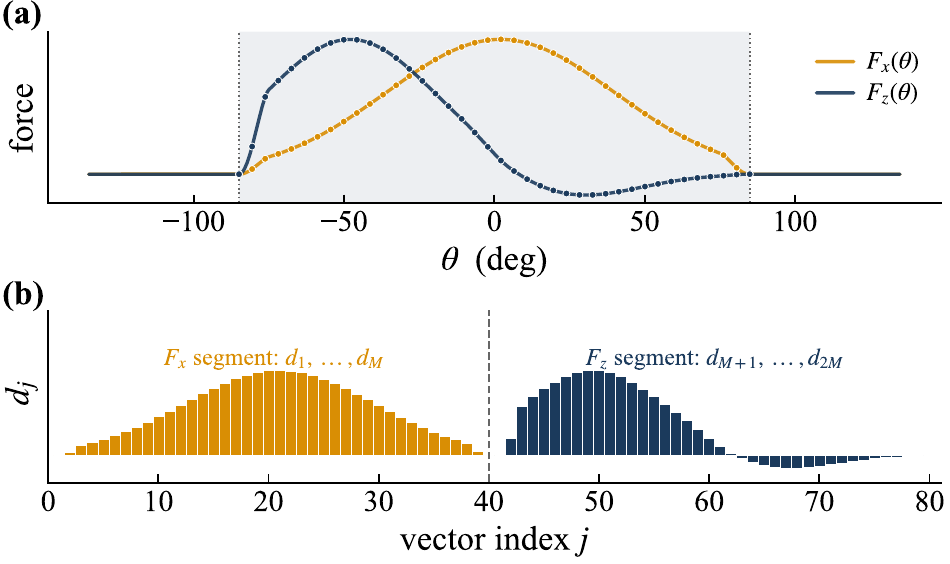}
  \caption{Observation vector $\mathbf{d}$: (a)~force histories resampled at
    40 angles over $|\theta|<85^\circ$; (b)~concatenated $F_x$ and $F_z$
    segments.}
  \label{fig:obs-vector}
\end{figure}

\begin{figure}[!htbp]
  \centering
  \includegraphics[width=0.75\linewidth]{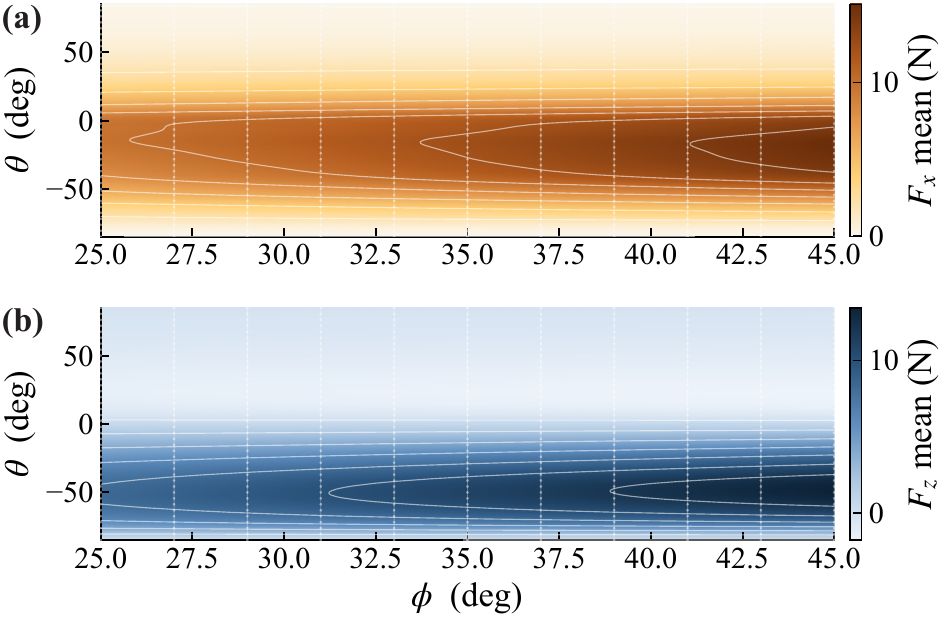}
  \caption{GP forward map: fitted mean surfaces of (a)~$F_x(\phi,\theta)$, and
    (b)~$F_z(\phi,\theta)$.}
  \label{fig:gp-mean-map}
\end{figure}

\begin{figure}[!htbp]
  \centering
  \includegraphics[width=0.75\linewidth]{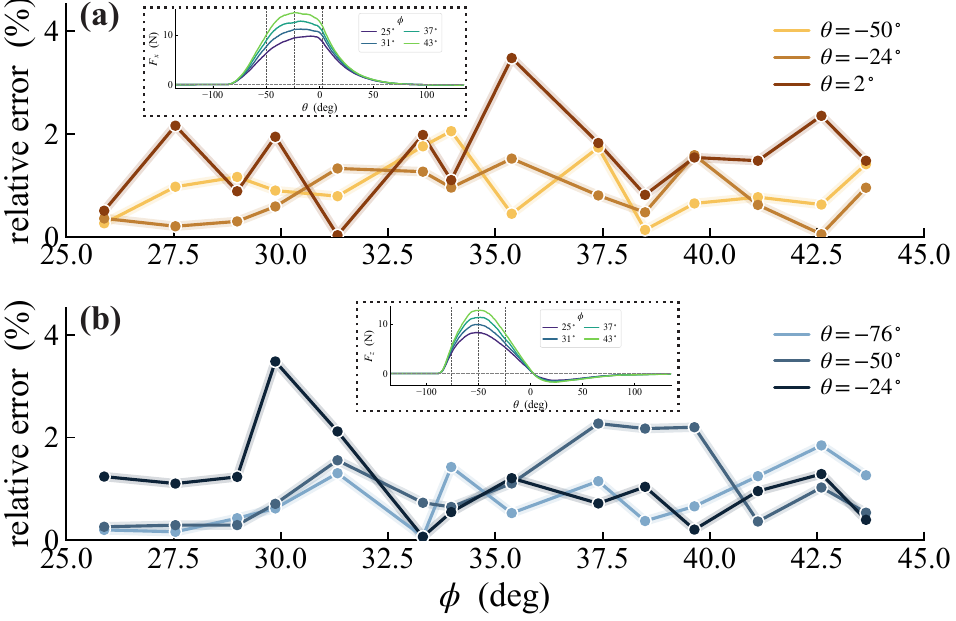}
  \caption{GP forward-map held-out relative error,
    $|F^{\mathrm{GP}}-F^{\mathrm{true}}|/F^{\mathrm{true}}$, between the GP
    mean and the true simulated value for each of the $14$ off-grid test cases
    described in Section~\ref{sec:inverse-synth}, evaluated at representative
    leg angles: (a)~$F_x$ and (b)~$F_z$. The insets show representative
    simulated force histories over the full leg-angle range, with vertical
    dashed lines indicating the angles selected for error evaluation.}
  \label{fig:gp-error-eval}
\end{figure}

\paragraph{Forward map}
Generating the simulation library of Section~\ref{sec:forward} with the forward operator is feasible once, but running a new simulation for every candidate $\phi$ during inversion would make repeated posterior evaluation prohibitively expensive. Because the force response varies smoothly with $\phi$ and $\theta$ (Fig.~\ref{fig:phi-sensitivity}), a Gaussian-process (GP) surrogate interpolates the simulated histories efficiently, following established use of response-surface and reduced-order models in geotechnical inverse analysis \cite{zhang2018determination,zhao2024uncertainty}. Each sweep is reduced to a fixed observation vector by resampling the thrust and lift histories onto a common grid of $M=40$ leg angles $\{\theta_j\}$ across the engaged window $|\theta|<85^\circ$, $\mathbf{d}=[F_x(\theta_1),\dots,F_x(\theta_M),\,F_z(\theta_1),\dots,F_z(\theta_M)]$ (Fig.~\ref{fig:obs-vector}). The grid is chosen dense enough to retain the full shape of the force history, including the descending and withdrawal portions that carry $\phi$-information beyond the peaks, rather than reducing each sweep to scalar features such as peak force. At the fixed benchmark rotation rate $\omega=0.2~\mathrm{rad\,s^{-1}}$, the prediction $\mathbf{g}(\phi)$ is formed by two GPs, one each for the
$F_x(\phi,\theta)$ and $F_z(\phi,\theta)$ surfaces, trained on all $11\times M$
(library friction angle, angle) pairs (Table~\ref{tab:library}) with an
anisotropic squared-exponential kernel and a white-noise term; hyperparameters
are fit by marginal-likelihood optimization. Inference is restricted to the
library span $\phi\in[25^\circ,45^\circ]$. Fig.~\ref{fig:gp-mean-map} shows the fitted mean surfaces, and
Fig.~\ref{fig:gp-error-eval} evaluates the resulting held-out error on the
14 off-grid test cases of Section~\ref{sec:inverse-synth} at three
representative leg angles.

\paragraph{Likelihood and residual covariance}
With residual $\mathbf{r}=\mathbf{d}-\mathbf{g}(\phi)$, the data are modeled as
zero-mean Gaussian \cite{stuart2010inverse},
\begin{equation}
  p(\mathbf{d}\mid\phi,s) = \mathcal{N}\!\big(\mathbf{r};\;
  \mathbf{0},\; s^{2}\boldsymbol{\Sigma}_{\mathrm{model}}\big),
  \label{eq:like}
\end{equation}
where $\boldsymbol{\Sigma}_{\mathrm{model}}$ represents GP interpolation error
and particle-sampling scatter. Their smooth variation with leg angle is modeled
by a squared-exponential covariance within each force component \cite{Rasmussen2006},
\begin{equation}
  [\boldsymbol{\Sigma}_{\mathrm{model}}]_{jk}
  = \sigma_g^{2}\,\exp\!\big[-(\theta_j-\theta_k)^{2}/2\ell^{2}\big],
  \label{eq:sig-model}
\end{equation}
with $\sigma_g = 0.01\,\Delta F_{\mathrm{pp}}$ and $\ell \approx 6^\circ$, where
$\Delta F_{\mathrm{pp}}$ is the local peak-to-peak range and $\ell$ is obtained
from the residual autocorrelation of the library curves. The covariance is
block diagonal in $F_x$ and $F_z$. Because $\Delta F_{\mathrm{pp}}$ is computed
once from the observation, the covariance is independent of $\phi$ and its
log-determinant is constant. A scalar covariance multiplier $s$ is assigned the
scale-invariant prior $p(s)\propto1/s$ and marginalized analytically, giving
\begin{equation}
\begin{aligned}
  \log p(\mathbf{d}\mid\phi) &= \mathrm{const}
  - \tfrac{n}{2}\,\log\!\big(\mathbf{r}^{\!\top}\boldsymbol{\Sigma}_{\mathrm{model}}^{-1}\mathbf{r}\big), \\
  n &= 2M = 80,
\end{aligned}
  \label{eq:marglike}
\end{equation}
which reduces sensitivity to the overall covariance scale.
Fig.~\ref{fig:obs-vector} illustrates the assembly of $\mathbf{d}$ from the
resampled $F_x$ and $F_z$ histories.

\begin{figure}[!htbp]
  \centering
  \includegraphics[width=0.75\linewidth]{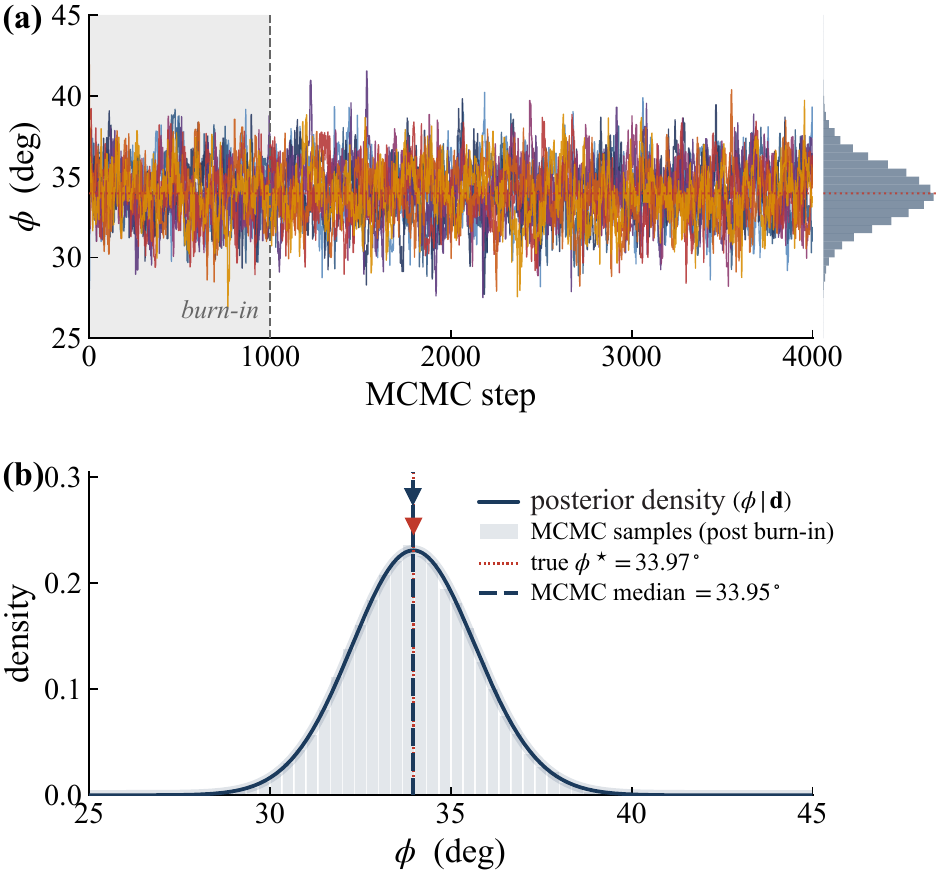}
  \caption{MCMC sampling for $\phi^\star=33.97^\circ$: (a)~walker traces and
    burn-in cutoff; (b)~posterior density and retained samples, with recovered $\phi=33.95^\circ$.}
  \label{fig:mcmc-illustration}
\end{figure}

\paragraph{Posterior sampling and reporting}
The prior is uniform over the library span, $\phi\sim\mathcal{U}(25^\circ,45^\circ)$. We sample each posterior (Eq.~\ref{eq:bayes} with Eq.~\ref{eq:marglike}) by Markov chain Monte Carlo (MCMC), which provides point estimates and credible intervals without imposing a Gaussian posterior and extends naturally to future multiparameter inference \cite{murakami2023developments}. We use a Goodman-Weare affine-invariant ensemble sampler~\cite{goodman2010ensemble} ($32$ walkers, $4\times10^3$ steps, walkers initialized uniformly over the prior, first $10^3$ steps discarded as burn-in), run independently for each of the 14 synthetic cases. Split-$\hat R$ and integrated-autocorrelation-time checks confirmed convergence for every posterior. Each posterior is summarized by its median and equal-tailed $68\%$ and $95\%$ credible intervals. Fig.~\ref{fig:mcmc-illustration} shows a representative case ($\phi^\star=33.97^\circ$): after burn-in the walker ensemble mixes stationarily about the true value (a), and the retained samples reproduce the likelihood map with an MCMC median of $33.95^\circ$, within $0.02^\circ$ of the truth and visually indistinguishable from it in the posterior density (b).

\subsection{Matched-model recovery experiments}
\label{sec:inverse-synth}
\begin{figure}[!htbp]
  \centering
  \includegraphics[width=0.75\linewidth]{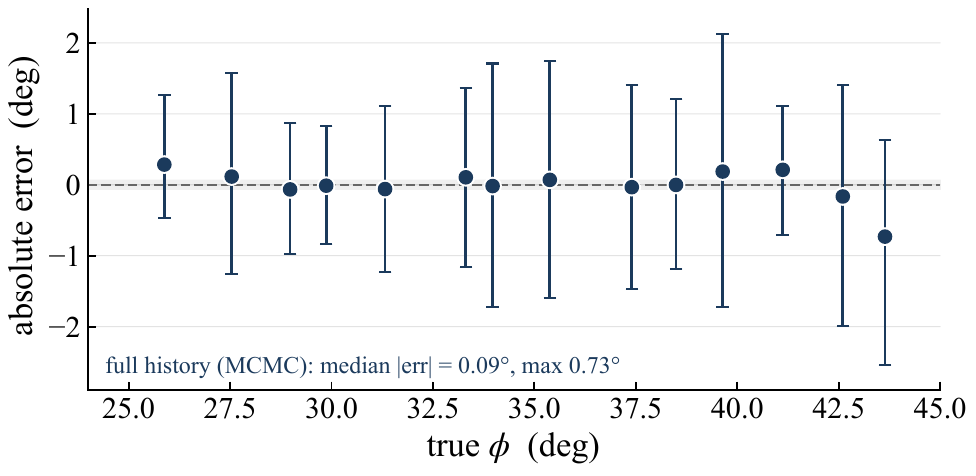}
  \caption{Matched-model recovery for 14 off-grid cases. Error bar of each recovered friction angle shows $68\%$
    credible intervals. Absolute
    error $=\phi_{\mathrm{recovered}}-\phi_{\mathrm{true}}$; the dashed
    line marks zero error.}
  \label{fig:phi-recovery}
\end{figure}

The matched-model behavior of the inversion is evaluated using fourteen friction angles $\phi^\star\in[25.9^\circ,43.6^\circ]$ placed off the library grid, with a fresh straight-leg simulation run for each at the benchmark rate $\omega=0.2\,\mathrm{rad\,s^{-1}}$. The truth runs and library
share the same solver and constitutive model, while the off-grid parameter
values provide an interpolation test. Each synthetic history is the clean
resampled truth, with no added measurement uncertainty, so the experiment
isolates interpolation and particle-sampling scatter alone. Fig.~\ref{fig:phi-recovery} shows the outcome. Using the joint-surface GP forward map and the full $M=40$-point history, all $14$ truths recover to within a maximum error of $0.73^\circ$ (median $0.09^\circ$). The $68\%$ credible intervals have a median half-width of $1.34^\circ$ (maximum $1.93^\circ$) and contain the true value in every case.

\FloatBarrier
\section{Discussion and conclusion}
\label{sec:discussion}

This study aims to test whether the internal friction angle of dry, cohesionless soil is identifiable from the force history of a rotating leg, and the framework answers this in the affirmative under matched modeling assumptions. A continuum forward model, benchmarked against measured rotating-leg force histories across three leg geometries, drives a pair of Gaussian-process surrogates and a Bayesian inversion of the full force trajectory. The simulated force history varies monotonically with friction angle, and the $\phi$-dependent information is distributed across the entire trajectory, including the descending and withdrawal portions, rather than confined to the peaks. Friction angle is recovered by inverting the full history rather than scalar features. In matched-model experiments, the framework recovers 14 off-grid friction angles with a median absolute error of approximately $0.1^\circ$ (maximum $\sim\!0.7^\circ$), and the reported $68\%$ credible intervals contain the truth in every case, so the inversion returns not only an accurate point estimate but a statement of uncertainty. Together these results establish conditional identifiability of $\phi$: the leg-force history carries a resolvable, quantitatively recoverable signature of soil friction angle when the forward model is correctly specified.
 
These results are obtained within a prescribed numerical model, so the posterior widths quantify identifiability rather than attainable physical accuracy. The plane-strain model, augmented by the out-of-plane relief formulation, reproduces the straight-leg benchmark within approximately $20\%$ while idealizing the three-dimensional geometry and trajectory of a quadruped foot, and the study is limited to homogeneous, dry cohesionless soil and one straight-leg geometry. Closing the gap to a field measurement will require extension to three dimensions, a foot- and gait-specific forward model, and validation against independent direct-shear or penetration ground truth, after which joint estimation of friction angle, density, and dilation from multiple maneuvers would move the method beyond the present one-parameter setting.

The field recordings confirm that the required force signal is available onboard; combined with the demonstrated identifiability of $\phi$, this supports proprioceptive soil sensing as a route to distributed, in-situ terrain characterization. Such measurements would feed the two applications that motivate this work: parameterizing physics-grounded world models with credible terrain mechanics for contact-rich robot training, and rapid slope-hazard screening in post-fire terrain.

\bibliographystyle{unsrt}
\bibliography{references}

\end{document}